\documentclass[runningheads,a4paper]{llncs}

\usepackage{graphicx}
\usepackage{amsmath,amssymb,amsfonts}

\usepackage{booktabs}
\usepackage{array}
\usepackage{multirow}
\usepackage[table]{xcolor}   
\usepackage{adjustbox}       

\usepackage[noadjust]{cite}

\usepackage{microtype}

\usepackage{url}
\usepackage[colorlinks,linkcolor=blue,citecolor=blue,urlcolor=blue]{hyperref}

\begin{document}

\title{NeuroWeaver: An Autonomous Evolutionary Agent for Exploring the Programmatic Space of EEG Analysis Pipelines}

\titlerunning{NeuroWeaver: An Autonomous Evolutionary Agent for EEG Pipelines}

\author{Guoan Wang \and Shihao Yang \and Feng Liu}
\authorrunning{G. Wang et al.}

\institute{Department of Industrial and Systems Engineering,\\
Rutgers, The State University of New Jersey,\\
96 Frelinghuysen Road, Piscataway, NJ 08854, USA\\
\email{feng.liu@rutgers.edu}}

\maketitle

\begin{abstract}
Although foundation models have achieved remarkable success in general domains, applying them to electroencephalography (EEG) analysis is constrained by substantial data requirements and large parameter counts, which incur prohibitive computational costs and impede deployment in resource-constrained clinical environments. General-purpose automated machine learning frameworks are likewise ill-suited to this domain, since exploration within an unbounded programmatic space fails to incorporate essential neurophysiological priors and frequently yields neuroscientifically implausible solutions. We therefore propose NeuroWeaver, a unified autonomous evolutionary agent that generalizes across diverse EEG datasets and tasks by reformulating pipeline engineering as a discrete constrained optimization problem solved through large language model (LLM)-driven generation of executable code. A Domain-Informed Subspace Initialization confines the search to a neuroscientifically plausible manifold, while a Multi-Objective Evolutionary Optimization dynamically balances performance, novelty, and efficiency via self-reflective refinement. Across five heterogeneous benchmarks, NeuroWeaver synthesizes lightweight pipelines that outperform state-of-the-art task-specific methods on nearly all metrics and attain accuracy comparable to large-scale foundation models, even surpassing them on the HMC and Workload benchmarks with only $0.18$M and $0.011$M parameters, respectively.

\end{abstract}

\section{Introduction}
Brain--computer interfaces (BCIs) establish direct communication channels between the brain and external devices, supporting applications such as neurofeedback therapy and assistive device control for individuals with severe motor impairments~\cite{mcfarland2011brain}. As a non-invasive, portable, and low-cost modality that records cortical activity at millisecond temporal resolution, electroencephalography (EEG) has become the most widely deployed substrate for clinical and cognitive applications~\cite{henry2006electroencephalography}. Realizing this potential ultimately depends on automated EEG analysis, which converts raw multichannel recordings into reliable predictions for problems such as abnormality detection~\cite{von2017electroencephalographic}, cognitive workload assessment~\cite{zyma2019electroencephalograms}, and sleep staging~\cite{alvarez2021inter}.

\begin{figure}[htbp]
\centering
\includegraphics[width=0.88\textwidth]{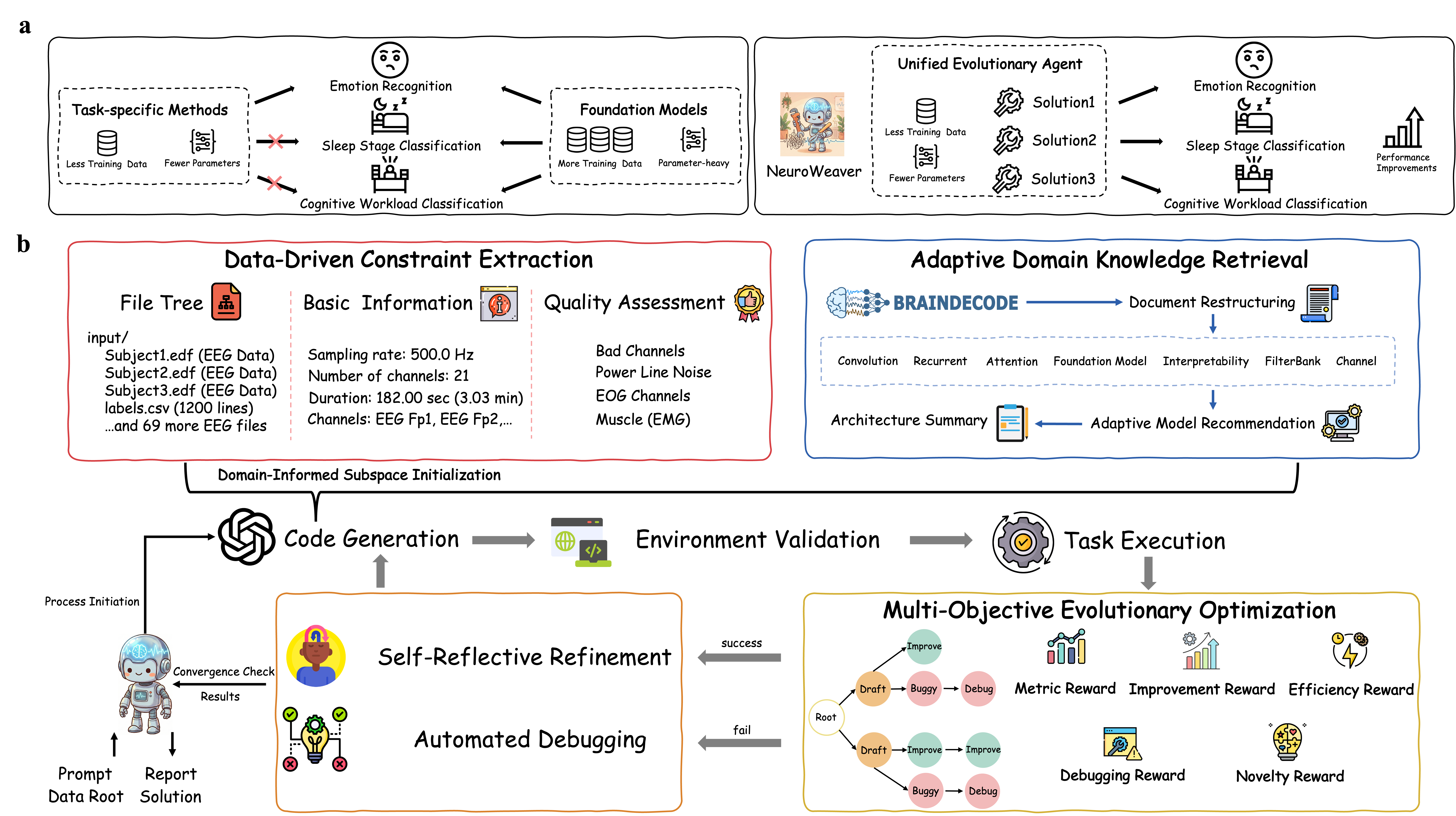}
\caption{\textbf{Framework overview of NeuroWeaver.} (a) Paradigm comparison, from static task-specific architectures and resource-intensive foundation models to a unified framework that autonomously synthesizes lightweight programmatic solutions. (b) The autonomous closed-loop workflow driven by Domain-Informed Subspace Initialization and Multi-Objective Evolutionary Optimization.} \label{fig:overview}
\end{figure}

Methodologically, EEG analysis has progressed from hand-crafted spectral, temporal, and connectivity descriptors paired with classical estimators~\cite{henry2006electroencephalography,harati2015improved} to end-to-end deep architectures that decode raw multichannel signals without manual feature design, including the compact convolutional network EEGNet~\cite{lawhern2018eegnet}, deeper convolutional models~\cite{schirrmeister2017deep,jing2023development}, contrastively trained encoders~\cite{yang2023self}, and recurrent or Transformer-based decoders~\cite{song2021transformer,peh2022transformer,li2022motor}. These supervised systems nevertheless remain inherently task-specific, since a separate model must be laboriously engineered by domain experts and trained from scratch for each downstream task, while pronounced data heterogeneity arising from variations in acquisition hardware, electrode configurations, and subject populations further undermines cross-dataset generalization~\cite{xu2020cross}.

To mitigate these limitations, recent research has gravitated toward EEG foundation models, which leverage large-scale self-supervised pretraining on unlabeled corpora to learn universal representations that are subsequently adapted to downstream tasks through supervised fine-tuning. BENDR transfers contrastive predictive coding from speech to scalp EEG~\cite{kostas2021bendr}, BIOT tokenizes each channel independently to support joint pretraining across mismatched montages and sampling rates~\cite{yang2023biot}, LaBraM casts pretraining as masked prediction over vector-quantized neural spectrum tokens~\cite{jiang2024large}, and NeuroLM aligns discrete EEG tokens with a language-model backbone so that a single universal model handles multiple downstream tasks at once~\cite{jiangneurolm}, while a rapidly growing body of work continues to extend this line~\cite{wang2024eegpt,el2026reve}. Although this paradigm substantially improves generalization, the parameter scale of these models renders downstream fine-tuning computationally expensive and inference cost substantial. Such costs are difficult to absorb in resource-constrained clinical environments and effectively preclude on-device deployment for many edge BCI scenarios.

In parallel with these developments, autonomous agents powered by large language models (LLMs) have emerged as a powerful paradigm for automated problem-solving~\cite{yao2023react}. In machine learning engineering, agents such as AIDE~\cite{jiang2025aide} and ML-Master~\cite{liu2025ml} cast solution discovery as a search over the programmatic space, combining tree-structured exploration with iterative refinement and demonstrating that LLM-driven agents can autonomously draft, execute, debug, and optimize pipelines without human intervention. These frameworks are nevertheless designed for general-purpose machine learning engineering and cannot be applied directly to EEG analysis: they operate without EEG-specific neurophysiological priors, and the pipelines that they synthesize are frequently non-executable in practice, stalling the iterative search at unsalvageable nodes. Within EEG analysis, this paradigm remains largely unexplored. EEGAgent~\cite{zhao2026eeg} operates as a scheduler over a fixed library of predefined tools and therefore lacks code-level optimization, with the scope further confined to a narrow set of preset tasks, while EEG-GPT~\cite{kim2024eeg} relies on fine-tuning an LLM and a single direct invocation to classify EEG, and thus constitutes a trained classifier rather than an autonomous agent system. The field thus still lacks a unified agent that can autonomously generate high-performance, lightweight, and task-specific analysis pipelines for arbitrary downstream EEG tasks.

To bridge this gap, we introduce \textbf{NeuroWeaver}
, an autonomous evolutionary agent that automates the engineering of EEG analysis pipelines by iteratively drafting, debugging, and optimizing programmatic solutions until a multi-objective reward is maximized. As illustrated in Fig.~\ref{fig:overview}a, NeuroWeaver functions as a \textbf{unified framework} capable of adapting to \textbf{arbitrary datasets and tasks}. By synthesizing distinct, task-specific code without labor-intensive expert engineering while autonomously orchestrating the entire lifecycle from model training to result analysis, NeuroWeaver produces analysis pipelines whose parameter counts are far smaller than those of foundation models, thereby enabling faster training, faster inference, and substantially easier deployment on edge devices.

Our contributions are summarized as follows. First, we propose the first unified end-to-end autonomous agent framework for automated EEG analysis pipeline engineering, which orchestrates the entire development lifecycle, including automated debugging and self-reflective refinement, without human intervention. Second, we introduce a domain-informed subspace initialization strategy that confines the search to a neuroscientifically plausible manifold, together with a multi-objective reward mechanism that dynamically balances accuracy, novelty, and computational cost. Third, extensive evaluations on five diverse benchmarks demonstrate that NeuroWeaver surpasses state-of-the-art task-specific methods across nearly all metrics with a minimal computational footprint, while also outperforming foundation models on the HMC~\cite{alvarez2021inter} and Workload~\cite{zyma2019electroencephalograms} datasets.

\section{Method}
\subsection{Problem Formulation}
We cast the automated design of EEG analysis pipelines as a discrete optimization problem over a programmatic space $\mathcal{S}$, defined as the set of all end-to-end executable scripts that map raw EEG recordings to task-specific predictions. Each candidate $s \in \mathcal{S}$ couples a three-stage prediction map with an evaluation module that scores its output:
\begin{equation}
\hat{y} = (f_{model} \circ f_{pre} \circ f_{load})(x), \qquad m_{s} = f_{eval}(\hat{y}, y)
\end{equation}
where $f_{load}(\cdot)$ standardizes the heterogeneous raw signals into a unified tensor format, $f_{pre}(\cdot)$ performs signal preprocessing such as spectral filtering and artifact removal, $f_{model}(\cdot)$ denotes the learning architecture together with the training procedure that fits its parameters, and the terminal module $f_{eval}(\cdot,\cdot)$ scores the predictions against the reference labels $y$ under the task metric, returning the scalar performance $m_{s}$. The objective is to identify the optimal script $s^*$ maximizing a composite utility $\mathcal{R}$ on an unseen test set $\mathcal{D}_{test}$, subject to the constraint that $s$ remains executable and logically consistent:
\begin{equation}
    s^* = \mathop{\mathrm{argmax}}_{s \in \mathcal{S}} \mathcal{R}(s \mid \mathcal{D}_{test})
\end{equation}
The utility $\mathcal{R}$ is anchored on $m_{s}$ and augments it with complementary objectives, formalized in Sec.~\ref{sec:moeo}. Since each candidate must be executable with mutually compatible modules and its quality emerges only after the code is run, navigating $\mathcal{S}$ is at once constrained and costly, which motivates the two components developed below.

\subsection{Overview of NeuroWeaver}
To explore the programmatic space $\mathcal{S}$, we instantiate the closed-loop workflow depicted in Fig.~\ref{fig:overview}b. Given a task description, evaluation criteria, and the raw data directory as inputs, the workflow begins with Data-Driven Constraint Extraction and Adaptive Domain Knowledge Retrieval, which together constitute the Domain-Informed Subspace Initialization (DISI): the agent identifies the intrinsic signal constraints of the dataset and retrieves neuro-architectural priors from a curated open-source library, restricting subsequent code generation to a neuroscientifically plausible sub-manifold of $\mathcal{S}$. Conditioned on this subspace, the agent leverages the generative coding capabilities of an LLM to synthesize candidate scripts, each of which undergoes Environment Validation to resolve dependencies before Task Execution. The resulting performance metrics, runtime statistics, and execution logs are then routed into Multi-Objective Evolutionary Optimization (MOEO), which governs the global search by maintaining a solution tree in which each node corresponds to a fully executable pipeline, until the convergence criteria are met and the agent returns the optimal pipeline $s^*$ with a summary report.

\subsection{Domain-Informed Subspace Initialization}
Generic LLMs generate code within the unbounded space $\mathcal{S}$, frequently yielding solutions that are syntactically valid yet neuroscientifically implausible. We therefore restrict the search to a domain-specific subspace $\mathcal{S}_{EEG} \subset \mathcal{S}$ through a structured initialization pipeline.

\noindent
\textbf{Data-Driven Constraint Extraction.}
An automated data preview routine analyzes the raw data directory and constructs a metadata descriptor $\phi_{data} = \langle \mathbf{i}_{EEG}, \mathbf{q}_{artifact} \rangle$, formalized as the extraction of a physical constraint vector rather than as mere file reading. The first component $\mathbf{i}_{EEG}$ summarizes the intrinsic acquisition attributes of the recording, namely the sampling rate, the montage channels, the per-recording duration, and the file format, which jointly determine admissible filter cutoffs, the temporal grid of epoching, and the shape of the input tensor that the downstream architecture must accept. The second component $\mathbf{q}_{artifact}$ aggregates rule-based diagnostics that quantify the dominant noise sources: spectral peaks at 50 or 60 Hz and their harmonics reveal the prevailing powerline frequency, channels matching canonical ocular nomenclature localize EOG contamination, and the proportion of power above 30 Hz flags channels with elevated EMG contamination. The descriptor thereby governs region-adaptive notch filtering, ocular artifact removal, and bad-channel handling, while preventing the input-to-architecture discrepancies that are typical of unconstrained code generation.

\noindent
\textbf{Adaptive Domain Knowledge Retrieval.}
To incorporate neuro-architectural priors, we design a two-stage retrieval mechanism over Braindecode~\cite{schirrmeister2017deep}, the most widely adopted open-source library for deep-learning analysis of EEG signals. First, given the objective of the downstream task $T_{goal}$, an LLM-based recommender ranks every entry of the model zoo against $T_{goal}$ and the evaluation criteria and selects one representative model $m_c$ for each category $c \in \mathcal{C}_{BD}$ of the official taxonomy of the library, which spans Convolution, Recurrent, Attention/Transformer, FilterBank, Channel, Foundation Model, and Interpretability. Second, to bridge the semantic gap between complex source code and LLM comprehension, a summarization function parses the source code and documentation of each $m_c$ and condenses the architectural primitives that distinguish it into a concise textual digest. The aggregated digests constitute the textual prior $\mathcal{T}_{prior}$, which provides a category-balanced view of proven design choices rather than a single canonical baseline.

\noindent
\textbf{Initialization of the Search Subspace.}
These two modalities jointly constrain the optimization problem, so that the agent generates the initial root solution $s_0$ via a conditional mapping rather than random initialization:
\begin{equation}
    s_0 = \pi_{\theta}\left( s \mid T_{goal}, \phi_{data}, \mathcal{T}_{prior} \right)
\end{equation}
By conditioning the generative policy $\pi_\theta$ on $\phi_{data}$ and $\mathcal{T}_{prior}$, every candidate script inherits a data-ingestion module aligned with the physical reality of EEG acquisition and a modeling module instantiating an architecture empirically validated by the EEG community. The search for $s^*$ is thereby confined to $\mathcal{S}_{EEG}$, in which vast regions that are syntactically reachable yet neuroscientifically implausible are pruned before iterative refinement begins.

\subsection{Multi-Objective Evolutionary Optimization}\label{sec:moeo}
Building on the tree-structured exploration paradigm of ML-Master~\cite{liu2025ml}, MOEO selects the next node to expand by descending the solution tree to the child maximizing the Upper Confidence Bounds applied to Trees criterion $\bar{R}_c + C\sqrt{\ln N_p / N_c}$, where $\bar{R}_c$ denotes the mean composite reward observed at the child and $N_c$ and $N_p$ are the visit counts of the child and its parent. The exploration constant $C$ decays piecewise from $\sqrt{2}$ toward $0.5$ across the iteration budget, so that early iterations favor breadth-oriented exploration and later iterations favor exploitation along the most promising lineages. Once a node is selected, the expansion routine is dispatched according to its execution status. A functional candidate triggers Self-Reflective Refinement, in which a diagnosis module synthesizes the architectural logic and quantitative outcomes of the parent into actionable instructions and prompts the agent to propose a single atomic modification, such as an adjustment to the preprocessing, the model architecture, or the training schedule. An erroneous candidate instead activates Automated Debugging, in which the same module distills the runtime traceback together with the offending source code, infers the root cause, and directs the agent to emit a corrected variant. The resulting child is retained in the tree but designated as the new local best of its lineage only if its primary metric strictly improves upon the strongest node observed so far on that branch, and its reward is then back-propagated through the lineage to update the value estimates that condition future selections.

To discourage premature convergence to local optima, the reward driving this process is itself multi-objective:
\begin{equation}
    R(s) = w_{m}\mathcal{M}(s|\mathcal{D}) + w_{i}\Delta(s, s_{parent}) + w_{n}\Omega(s) + w_{e}\Gamma(\tau_{s}) + w_{fix}\Phi(s)
\end{equation}
where each term is mapped to $[0,1]$ and the coefficients control the trade-off among five complementary objectives. The accuracy term $\mathcal{M}(s\mid\mathcal{D})$ applies min--max normalization to the primary metric using the running extrema observed across all evaluated candidates, yielding a signal comparable across heterogeneous benchmarks. The improvement term $\Delta(s, s_{parent})$ measures the gain of $s$ over the running best on the same lineage and clips non-positive values to zero, rewarding strict advancement rather than replication of an already-strong parent. The novelty term $\Omega(s)$ is supplied by an LLM-based judge that compares $s$ against a buffer of recently evaluated candidates along seven dimensions, spanning model architecture, preprocessing, training strategy, regularization, signal-processing choices, evaluation protocol, and implementation pattern. The efficiency term $\Gamma(\tau_{s}) = 1 - \sqrt{\min(1, \tau_{s}/\tau_{max})}$, with $\tau_{s}$ the execution latency of $s$ and $\tau_{max}$ the timeout budget, applies a smooth penalty rather than a hard cutoff, so that low-latency solutions are preferred at comparable accuracy. Finally, the robustness term $\Phi(s)$ awards a bug-fix bonus when $s$ repairs a buggy parent, channeling search effort toward recovering failing branches rather than abandoning them. Candidates that fail to execute bypass this formulation and receive a flat penalty $R(s)=-1$, so that selection demotes infeasible regions of the search space.

\section{Experiments}

\begin{table}[htbp]
\centering
\caption{\textbf{Summary of the five downstream EEG benchmarks.} For each dataset we list the channel count, sampling rate, epoch duration, and the target classification task with the corresponding number of classes.}\label{tab:datasets}
\fontsize{8}{9}\selectfont
\setlength{\tabcolsep}{4pt}
\renewcommand{\arraystretch}{1.1}
\begin{adjustbox}{max width=\textwidth}
\begin{tabular}{lcccc}
\toprule
\textbf{Dataset} & \textbf{Channel} & \textbf{Sampling Rate} & \textbf{Duration} & \textbf{Task} \\
\midrule
TUEV~\cite{harati2015improved}     & 23 & 256 Hz  & 5 seconds  & event type classification (6-class)         \\
SEED~\cite{zheng2015investigating} & 62 & 1000 Hz & 4 seconds  & emotion recognition (3-class)               \\
HMC~\cite{alvarez2021inter}        & 4  & 256 Hz  & 30 seconds & sleep stage classification (5-class)        \\
Workload~\cite{zyma2019electroencephalograms} & 19 & 500 Hz  & 4 seconds  & cognitive workload classification (Binary)  \\
TUSL~\cite{von2017electroencephalographic}    & 23 & 256 Hz  & 10 seconds & slowing event classification (3-class)      \\
\bottomrule
\end{tabular}
\end{adjustbox}
\end{table}

\subsection{Experimental Setup}
To ensure a rigorous and fair comparison with state-of-the-art methods, we align our protocols with NeuroLM~\cite{jiangneurolm}. We evaluate NeuroWeaver on five heterogeneous benchmarks (Table~\ref{tab:datasets})---TUEV~\cite{harati2015improved}, SEED~\cite{zheng2015investigating}, HMC~\cite{alvarez2021inter}, Workload~\cite{zyma2019electroencephalograms}, and TUSL~\cite{von2017electroencephalographic}---encompassing diverse cognitive and physiological tasks. To prevent leakage, data partitioning follows the subject-independent, chronological, or random strategies defined by the baselines. Preprocessing is standardized across all compared methods: raw signals undergo 0.1--75 Hz bandpass filtering, region-adaptive notch filtering (50/60 Hz), resampling to 200 Hz, and global amplitude scaling by $10^{-2}$; no constraint is imposed on any subsequent processing that NeuroWeaver introduces into the synthesized pipeline. We report Balanced Accuracy as the primary metric, together with Cohen $\kappa$ and Weighted F1-score for multi-class tasks and AUROC and AUC-PR for binary classification. The framework employs ChatGPT-5.1~\cite{achiam2023gpt} as the primary backbone, complemented by ChatGPT-4.1 for self-reflective refinement and ChatGPT-4o for novelty scoring. Evolutionary optimization operates with a global budget of 200 iterations and a parallelism degree of 3, incurring an approximate API cost of \$20 per run. The composite reward is instantiated with $w_{m}=1.0$, $w_{i}=0.5$, $w_{n}=0.3$, $w_{e}=0.1$, and $w_{fix}=0.5$; each search begins with five root drafts and admits at most three improvement children per executable parent, and the UCT exploration constant decays piecewise from $\sqrt{2}$ to $0.5$ between the $30\%$ and $70\%$ marks of the budget. The complete search configuration is released with the source code.

\begin{table}[htbp]
\centering
\caption{\textbf{Performance comparison across the five downstream benchmarks.} Balanced Accuracy, Cohen $\kappa$, and Weighted F1 are reported for the multi-class tasks and Balanced Accuracy, AUC-PR, and AUROC for the binary Workload task, as mean $\pm$ standard deviation over three independent runs. \underline{Underlined} entries mark the best result within each method category, while \textcolor{blue}{blue} indicates that NeuroWeaver surpasses the strongest task-specific baseline. Paired model sizes correspond to the two datasets of the respective panel.}\label{tab2}
\fontsize{8}{9}\selectfont
\setlength{\tabcolsep}{3pt}
\renewcommand{\arraystretch}{1.05}
\begin{adjustbox}{max width=\textwidth}
\begin{tabular}{lc ccc ccc}
\toprule
\multirow{2}{*}{Method} & \multirow{2}{*}{Model Size} &
\multicolumn{3}{c}{SEED} & \multicolumn{3}{c}{HMC} \\
\cmidrule(lr){3-5}\cmidrule(lr){6-8}
& & Balanced Acc. & Cohen's Kappa & Weighted F1
  & Balanced Acc. & Cohen's Kappa & Weighted F1 \\
\midrule
\rowcolor{gray!15}\multicolumn{8}{c}{\textit{Task-specific Methods}} \\
SPaRCNet~\cite{jing2023development} & 0.79M & 0.5596 $\pm$ 0.0244 & 0.3464 $\pm$ 0.0372 & 0.5585 $\pm$ 0.0297 & 0.4756 $\pm$ 0.1109 & 0.3147 $\pm$ 0.1315 & 0.4108 $\pm$ 0.1310 \\
ContraWR~\cite{yang2023self} & 1.6M  & 0.6106 $\pm$ 0.0078 & 0.4220 $\pm$ 0.0129 & 0.6137 $\pm$ 0.0085 & 0.4242 $\pm$ 0.0541 & 0.2340 $\pm$ 0.0554 & 0.2987 $\pm$ 0.0288 \\
CNN-Transformer~\cite{peh2022transformer} & 3.2M & \underline{0.6161} $\pm$ 0.0384 & \underline{0.4262} $\pm$ 0.0601 & \underline{0.6150} $\pm$ 0.0463 & \underline{0.6573} $\pm$ 0.0141 & \underline{0.5961} $\pm$ 0.0105 & \underline{0.6896} $\pm$ 0.0065 \\
FFCL~\cite{li2022motor} & 2.4M & 0.5808 $\pm$ 0.0322 & 0.3732 $\pm$ 0.0462 & 0.5743 $\pm$ 0.0402 & 0.4427 $\pm$ 0.0702 & 0.2542 $\pm$ 0.0654 & 0.2902 $\pm$ 0.0485 \\
ST-Transformer~\cite{song2021transformer} & 3.5M & 0.5479 $\pm$ 0.0091 & 0.3261 $\pm$ 0.0169 & 0.5505 $\pm$ 0.0091 & 0.2559 $\pm$ 0.0141 & 0.0503 $\pm$ 0.0183 & 0.1428 $\pm$ 0.0122 \\
\rowcolor{gray!15}\multicolumn{8}{c}{\textit{Pretrained Foundation Models}} \\
BIOT~\cite{yang2023biot} & 3.2M & 0.7097 $\pm$ 0.0024 & 0.5682 $\pm$ 0.0051 & 0.7134 $\pm$ 0.0027 & 0.6862 $\pm$ 0.0041 & 0.6295 $\pm$ 0.0113 & 0.7091 $\pm$ 0.0147 \\
LaBraM-Base~\cite{jiang2024large} & 5.8M & \underline{0.7318} $\pm$ 0.0019 & \underline{0.5994} $\pm$ 0.0031 & \underline{0.7354} $\pm$ 0.0021 & \underline{0.7286} $\pm$ 0.0101 & \underline{0.6812} $\pm$ 0.0073 & \underline{0.7554} $\pm$ 0.0024 \\
NeuroLM-B~\cite{jiangneurolm} & 254M & 0.5554 $\pm$ 0.0075 & 0.3393 $\pm$ 0.0117 & 0.5599 $\pm$ 0.0068 & 0.6737 $\pm$ 0.0050 & 0.6188 $\pm$ 0.0057 & 0.7126 $\pm$ 0.0034 \\
\rowcolor{gray!15}\multicolumn{8}{c}{\textit{Agent System (Ours)}} \\
NeuroWeaver & 0.37M/0.18M   & \textcolor{blue}{0.6211} $\pm$ 0.0058 & \textcolor{blue}{0.4397} $\pm$ 0.0087 & \textcolor{blue}{0.6246} $\pm$ 0.0058 & \textcolor{blue}{0.7862} $\pm$ 0.0006 & \textcolor{blue}{0.7160} $\pm$ 0.0003 & \textcolor{blue}{0.7794} $\pm$ 0.0004 \\
\midrule[\heavyrulewidth]
\multirow{2}{*}{Method} & \multirow{2}{*}{Model Size} &
\multicolumn{3}{c}{TUEV} & \multicolumn{3}{c}{TUSL} \\
\cmidrule(lr){3-5}\cmidrule(lr){6-8}
& & Balanced Acc. & Cohen's Kappa & Weighted F1
  & Balanced Acc. & Cohen's Kappa & Weighted F1 \\
\midrule
\rowcolor{gray!15}\multicolumn{8}{c}{\textit{Task-specific Methods}} \\
SPaRCNet~\cite{jing2023development} & 0.79M & 0.4161 $\pm$ 0.0262 & \underline{0.4233} $\pm$ 0.0181 & \underline{0.7024} $\pm$ 0.0104 & 0.4185 $\pm$ 0.0452 & 0.1399 $\pm$ 0.0799 & 0.3500 $\pm$ 0.0968 \\
ContraWR~\cite{yang2023self} & 1.6M & \underline{0.4384} $\pm$ 0.0349 & 0.3912 $\pm$ 0.0237 & 0.6893 $\pm$ 0.0136 & \underline{0.5857} $\pm$ 0.0662 & \underline{0.3567} $\pm$ 0.0968 & \underline{0.5458} $\pm$ 0.0798 \\
CNN-Transformer~\cite{peh2022transformer} & 3.2M & 0.4087 $\pm$ 0.0161 & 0.3815 $\pm$ 0.0134 & 0.6854 $\pm$ 0.0293 & 0.3575 $\pm$ 0.0151 & 0.0306 $\pm$ 0.0179 & 0.2235 $\pm$ 0.0251 \\
FFCL~\cite{li2022motor} & 2.4M & 0.3979 $\pm$ 0.0104 & 0.3732 $\pm$ 0.0188 & 0.6783 $\pm$ 0.0120 & 0.3819 $\pm$ 0.0688 & 0.0628 $\pm$ 0.0888 & 0.2120 $\pm$ 0.0786 \\
ST-Transformer~\cite{song2021transformer} & 3.5M & 0.3984 $\pm$ 0.0228 & 0.3765 $\pm$ 0.0306 & 0.6823 $\pm$ 0.0190 & 0.4000 $\pm$ 0.0329 & 0.0860 $\pm$ 0.0449 & 0.3793 $\pm$ 0.0459 \\
\rowcolor{gray!15}\multicolumn{8}{c}{\textit{Pretrained Foundation Models}} \\
BIOT~\cite{yang2023biot} & 3.2M & 0.5281 $\pm$ 0.0225 & 0.5273 $\pm$ 0.0249 & 0.7492 $\pm$ 0.0082 & 0.5758 $\pm$ 0.0303 & 0.2012 $\pm$ 0.0212 & 0.2394 $\pm$ 0.0040 \\
LaBraM-Base~\cite{jiang2024large} & 5.8M & \underline{0.6409} $\pm$ 0.0065 & \underline{0.6637} $\pm$ 0.0093 & \underline{0.8312} $\pm$ 0.0052 & \underline{0.7625} $\pm$ 0.0131 & \underline{0.6407} $\pm$ 0.0304 & \underline{0.7614} $\pm$ 0.0210 \\
NeuroLM-B~\cite{jiangneurolm} & 254M & 0.4560 $\pm$ 0.0048 & 0.4285 $\pm$ 0.0048 & 0.7153 $\pm$ 0.0028 & 0.6734 $\pm$ 0.0436 & 0.5107 $\pm$ 0.0617 & 0.6743 $\pm$ 0.0394 \\
\rowcolor{gray!15}\multicolumn{8}{c}{\textit{Agent System (Ours)}} \\
NeuroWeaver & 1.2M/0.46M & \textcolor{blue}{0.4887} $\pm$ 0.0057 & \textcolor{blue}{0.4400} $\pm$ 0.0203 & \textcolor{blue}{0.7205} $\pm$ 0.0163 & \textcolor{blue}{0.7138} $\pm$ 0.0117 & 0.2206 $\pm$ 0.0281 & 0.4373 $\pm$ 0.0329 \\
\midrule[\heavyrulewidth]
\multirow{2}{*}{Method} & \multirow{2}{*}{Model Size} & \multicolumn{6}{c}{Workload} \\
\cmidrule(lr){3-8}
& & \multicolumn{2}{c}{Balanced Acc.} & \multicolumn{2}{c}{AUC-PR} & \multicolumn{2}{c}{AUROC} \\
\midrule
\rowcolor{gray!15}\multicolumn{8}{c}{\textit{Task-specific Methods}} \\
SPaRCNet~\cite{jing2023development} & 0.79M & \multicolumn{2}{c}{0.5977 $\pm$ 0.0071} & \multicolumn{2}{c}{0.6638 $\pm$ 0.0314} & \multicolumn{2}{c}{0.6717 $\pm$ 0.0172} \\
ContraWR~\cite{yang2023self} & 1.6M & \multicolumn{2}{c}{0.6966 $\pm$ 0.0332} & \multicolumn{2}{c}{0.7668 $\pm$ 0.0408} & \multicolumn{2}{c}{0.7685 $\pm$ 0.0317} \\
CNN-Transformer~\cite{peh2022transformer} & 3.2M & \multicolumn{2}{c}{0.5793 $\pm$ 0.0230} & \multicolumn{2}{c}{0.5306 $\pm$ 0.0459} & \multicolumn{2}{c}{0.5663 $\pm$ 0.0349} \\
FFCL~\cite{li2022motor} & 2.4M & \multicolumn{2}{c}{\underline{0.7069} $\pm$ 0.0197} & \multicolumn{2}{c}{\underline{0.7823} $\pm$ 0.0099} & \multicolumn{2}{c}{\underline{0.7857} $\pm$ 0.0234} \\
ST-Transformer~\cite{song2021transformer} & 3.5M & \multicolumn{2}{c}{0.6103 $\pm$ 0.0056} & \multicolumn{2}{c}{0.5716 $\pm$ 0.0071} & \multicolumn{2}{c}{0.6375 $\pm$ 0.0078} \\
\rowcolor{gray!15}\multicolumn{8}{c}{\textit{Pretrained Foundation Models}} \\
BIOT~\cite{yang2023biot} & 3.2M & \multicolumn{2}{c}{\underline{0.6655} $\pm$ 0.0665} & \multicolumn{2}{c}{\underline{0.7189} $\pm$ 0.0722} & \multicolumn{2}{c}{\underline{0.7342} $\pm$ 0.0536} \\
LaBraM-Base~\cite{jiang2024large} & 5.8M & \multicolumn{2}{c}{0.6609 $\pm$ 0.0204} & \multicolumn{2}{c}{0.7174 $\pm$ 0.0234} & \multicolumn{2}{c}{0.7272 $\pm$ 0.0165} \\
NeuroLM-B~\cite{jiangneurolm} & 254M & \multicolumn{2}{c}{0.6172 $\pm$ 0.0113} & \multicolumn{2}{c}{0.5824 $\pm$ 0.0080} & \multicolumn{2}{c}{0.6253 $\pm$ 0.0160} \\
\rowcolor{gray!15}\multicolumn{8}{c}{\textit{Agent System (Ours)}} \\
NeuroWeaver & 0.011M & \multicolumn{2}{c}{\textcolor{blue}{0.7391} $\pm$ 0.0439} & \multicolumn{2}{c}{\textcolor{blue}{0.8436} $\pm$ 0.0373} & \multicolumn{2}{c}{\textcolor{blue}{0.8403} $\pm$ 0.0361} \\
\bottomrule
\end{tabular}
\end{adjustbox}
\end{table}

\subsection{Comparison with State-of-the-Art Methods}
We evaluate NeuroWeaver against state-of-the-art task-specific methods~\cite{jing2023development,yang2023self,peh2022transformer,li2022motor,song2021transformer} and large-scale pretrained foundation models~\cite{yang2023biot,jiang2024large,jiangneurolm} across the five datasets (Table~\ref{tab2}), selecting the optimal synthesized pipeline $s^*$ for each benchmark and reporting the mean and standard deviation over three independent trials with distinct seeds. NeuroWeaver surpasses the strongest task-specific baseline on nearly all metrics with only $0.011$--$1.2$M parameters, compared with $0.79$--$3.5$M for the task-specific baselines and $3.2$--$254$M for the foundation models. On SEED, TUEV, and TUSL, the $0.37$M-, $1.2$M-, and $0.46$M-parameter pipelines exceed every task-specific baseline on Balanced Accuracy while consuming a small fraction of the budget required by the foundation models; on HMC and Workload they further surpass every pretrained foundation model on all reported metrics, despite using only $0.18$M and $0.011$M parameters.

\subsection{Characterization of the Synthesized Pipelines}\label{sec:pipeline_char}

\begin{table}[htbp]
\centering
\caption{\textbf{Summary of the optimal pipelines $s^*$ synthesized by NeuroWeaver on the five benchmarks.} Each row reports the architecture, training and augmentation strategies, and parameter count.}\label{tab5}
\fontsize{8}{9}\selectfont
\setlength{\tabcolsep}{4pt}
\renewcommand{\arraystretch}{1.1}
\begin{adjustbox}{max width=\textwidth}
\begin{tabular}{l l l c}
\toprule
Dataset & Architecture & Training \& Augmentation & \# Params \\
\midrule
SEED     & Temporal 1D CNN + squeeze-and-excitation + attention pooling & Channel dropout, EMA, label smoothing & 0.37M \\
HMC      & CNN + 2-layer BiGRU + temporal attention    & Mixup, SpecAugment, OneCycleLR        & 0.18M \\
TUEV     & 4-stage ResNet-1D                           & Temporal jitter, cosine warm restarts  & 1.20M \\
TUSL     & 2D CNN + spatial dropout                    & Per-channel z-score, Gaussian noise, cosine LR & 0.46M \\
Workload & EEGNet~\cite{lawhern2018eegnet} + squeeze-and-excitation & Mixup, label smoothing, cosine LR      & 0.011M \\
\bottomrule
\end{tabular}
\end{adjustbox}
\end{table}

Table~\ref{tab5} reports the backbone, the principal training and augmentation choices, and the parameter count of $s^*$ on every benchmark, and three regularities emerge. First, the agent exploits sequential structure when the epoch is long and the label depends on slow temporal dynamics, as on the $30$-second epochs of HMC, and otherwise prefers convolutional backbones whose depth scales with the number of target classes, as on the six-class TUEV and the three-class SEED. Second, the synthesized parameter counts grow with task complexity rather than with raw input volume, since the binary Workload task absorbs the smallest budget and the six-class TUEV task the largest, indicating that the multi-objective reward penalizes capacity that the task does not consume. Third, the regularization recipes align with the dominant nuisance factor of each dataset, namely subject-level heterogeneity on HMC and Workload, trial-level acquisition noise on TUSL, and class-imbalance pressure on SEED and TUEV. The cases in which foundation models retain an edge correspond to channel-rich, semantically complex tasks, on which the cross-dataset representations acquired through large-scale pretraining confer an advantage that a per-dataset search training every candidate from scratch cannot fully reproduce.

\begin{figure}[htbp]
\centering
\includegraphics[width=\textwidth]{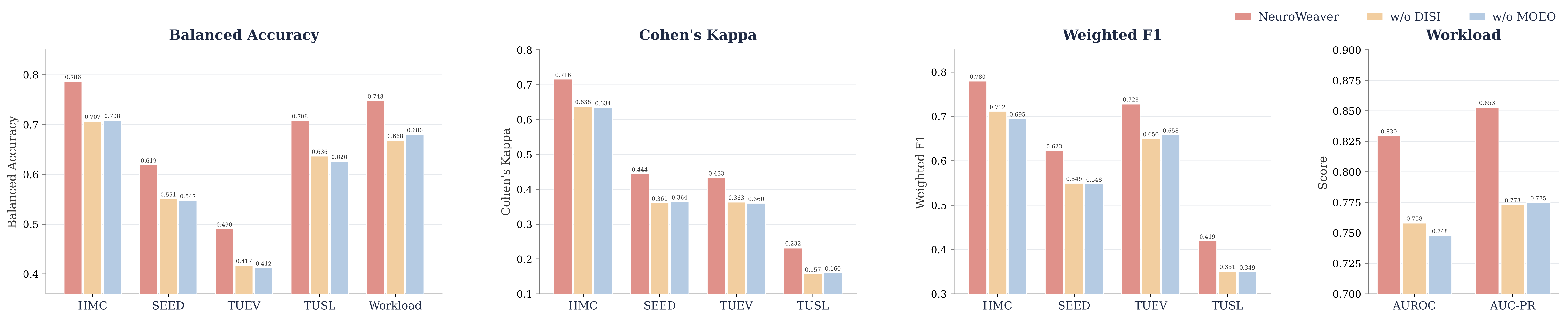}
\caption{\textbf{Ablation study of NeuroWeaver across the five benchmarks.} Grouped bars compare the full system with two variants that disable the Domain-Informed Subspace Initialization (w/o DISI) and the Multi-Objective Evolutionary Optimization (w/o MOEO).} \label{fig:abalation}
\end{figure}

\subsection{Ablation Study}
To evaluate the contribution of each design component, we ablate DISI and MOEO, reporting Balanced Accuracy, Cohen $\kappa$, and Weighted F1 on the multi-class benchmarks together with AUROC and AUC-PR on the binary Workload benchmark. Removing DISI reverts the agent to unconstrained generation over the full space $\mathcal{S}$, exposing the search to invalid input--architecture configurations and to model families poorly aligned with the characteristics of EEG, whereas removing MOEO disables the tree-structured search together with the composite reward. As shown in Fig.~\ref{fig:abalation}, both variants degrade substantially across all five benchmarks on primary and secondary metrics alike, confirming that the two components are complementary: DISI restricts the search to a neuroscientifically plausible subspace, whereas MOEO turns this subspace into a productive optimization landscape.

\section{Conclusion}
We have introduced NeuroWeaver, a unified autonomous evolutionary agent that reformulates EEG pipeline engineering as a domain-constrained search over the programmatic space and synthesizes a distinct, task-tailored analysis script for every dataset and downstream task. The framework addresses two long-standing limitations of contemporary EEG analysis: on the task-specific side it eliminates the labor-intensive practice in which domain experts hand-design, train, and tune a bespoke pipeline for every new dataset, and on the foundation-model side it removes the heavy parameter footprint together with the prohibitive fine-tuning and inference cost of large pretrained backbones. Evaluations across five heterogeneous benchmarks confirm that this single design synthesizes lightweight pipelines whose accuracy is competitive with foundation models orders of magnitude larger, and on HMC~\cite{alvarez2021inter} and Workload~\cite{zyma2019electroencephalograms} surpasses them.

A property that materially distinguishes NeuroWeaver from cloud-resident analytics services is the strictly local nature of every data-touching component. Data-Driven Constraint Extraction, Adaptive Domain Knowledge Retrieval, and the iterative code execution that grounds MOEO all run within the local runtime of the user, so the raw multichannel recordings never leave the host machine on which the data are stored. The interaction with external LLM providers is restricted to a small, desensitized set of metadata, namely the recording montage, the natural-language task and evaluation description, a compact set of acquisition attributes, and the scalar performance metrics of completed runs. By architecturally decoupling semantic reasoning from data execution, NeuroWeaver remains compatible with clinical infrastructures in which regulatory constraints forbid the disclosure of patient-identifiable neural data to third-party services.

Two limitations delimit the present study. First, the preprocessing pipeline was deliberately frozen to match the protocol of NeuroLM~\cite{jiangneurolm}; while this removes a major confounder, it contracts the optimization landscape, since signal cleaning and artifact rejection are precisely the dimensions on which a domain-aware agent is most likely to discover dataset-specific gains, and the descriptor $\phi_{data}$ is already well positioned to drive recording-adaptive preprocessing once the constraint is relaxed. Second, the agent inherits the code-synthesis capability and the stochastic decoding of its LLM backbone, so the quality and run-to-run consistency of the synthesized pipelines remain bounded by the present state of LLM technology; conversely, future advances in that technology will translate into commensurate gains without any architectural change. NeuroWeaver thereby establishes a viable foundation for an agentic mode of EEG analysis, in which producing a competitive pipeline for an arbitrary dataset and task no longer requires an expert-driven engineering cycle but unfolds as a constrained autonomous search conditioned on explicit neurophysiological priors.

\begin{credits}
\subsubsection{\discintname}
The authors have no competing interests to declare that are relevant to the content of this article.
\end{credits}

\bibliographystyle{splncs04}
\bibliography{citation}

\end{document}